\documentclass[preprintnumbers,showkeys,amsmath,amssymb]{revtex4}

\usepackage{graphicx}
\usepackage{dcolumn}
\usepackage{bm}

\begin{document}

\title{Decision Maker based on Atomic Switches}

\author{Song-Ju Kim$^{\dagger}$}
\email{KIM.Songju@nims.go.jp}

\author{Tohru Tsuruoka$^{\dagger}$}

\author{Tsuyoshi Hasegawa$^{\ddagger}$}

\author{Masakazu Aono$^{\dagger}$}

\affiliation{%
$\dagger$WPI Center for Materials Nanoarchitectonics, National Institute for Materials Science,\\ 
1--1 Namiki, Tsukuba, Ibaraki 305--0044, Japan}

\affiliation{%
$\ddagger$Department of Applied Physics, Waseda University,\\ 
3--4--1 Ookubo, Shinjuku-ku, Tokyo 169-8555, Japan
}

\begin{abstract}
We propose a simple model for an atomic switch-based decision maker (ASDM), and show that, as long as its total volume of precipitated Ag atoms is conserved when coupled with suitable operations, an atomic switch system provides a sophisticated ``decision-making'' capability that is known to be one of the most important intellectual abilities in human beings. 
We considered the multi-armed bandit problem (MAB); the problem of finding, as accurately and quickly as possible, the most profitable option from a set of options that gives stochastic rewards.
These decisions are made as dictated by each volume of precipitated Ag atoms, which is moved in a manner similar to the fluctuations of a rigid body in a tug-of-war game.
The ``tug-of-war (TOW) dynamics'' of the ASDM exhibits higher efficiency than conventional MAB solvers.
We show analytical calculations that validate the statistical reasons for the ASDM dynamics to produce such high performance, despite its simplicity.
These results imply that various physical systems, in which some conservation law holds, can be used to implement efficient ``decision-making objects.''
Efficient MAB solvers are useful for many practical applications, because MAB abstracts a variety of decision-making problems in real-world situations where an efficient trial-and-error is required.
The proposed scheme will introduce a new physics-based analog computing paradigm, which will include such things as ``intelligent nanodevices'' and ``intelligent information networks'' based on self-detection and self-judgment.
\end{abstract}


\maketitle

\section{Introduction}

When we look at the natural world, information processing in biological systems is elegantly coupled with their underlying physics~\cite{natcom,natcom2}.
This suggests a potential for establishing a new physics-based analog-computing paradigm.
A proposal was made about ten years ago for a conceptually novel switching device, called the ``atomic switch,'' that is based on metal ion migration and electrochemical reactions in solid electrolytes~\cite{terabe}.
Because its resistance state is controlled continuously by the movement of a limited number of metal ions/atoms, the atomic switch can be regarded as a physics-based analog-computing element.
In this paper, using atomic switches, we show that a physical constraint, the volume conservation law, allows for the efficient solving of decision-making problems which, in human beings, is one of the most important intellectual abilities.
 
Suppose there are $M$ slot machines, each of which returns a reward; for example, coins, with a certain probability density function (PDF) that is unknown to a player.
Let us consider a minimal case: two machines A and B give rewards with individual PDF whose mean reward is $\mu_A$ and $\mu_B$, respectively.
The player makes a decision on which machine to play at each trial, trying to maximize the total reward obtained after repeating several trials.
The multi-armed bandit problem (MAB) is used to determine the optimal strategy for playing machines as accurately and quickly as possible by referring to past experience.

In the context of decision making algorithms, the MAB was originally described by Robbins~\cite{robbins}, although the essence of the problem had been studied earlier by Thompson~\cite{thompson}.
The optimal strategy, called the ``Gittins index,'' is known only for a limited class of problems in which the reward distributions are assumed to be known to the players~\cite{gittins1,gittins2}.
Even in this limited class, in practice, computing the Gittins index becomes intractable for many cases.  
For the algorithms proposed by Agrawal and Auer et al., another index was expressed as a simple function of the reward sums obtained from the machines~\cite{agra,auer}.
In particular, the ``upper confidence bound 1 (UCB1) algorithm'' for solving MABs is used worldwide in many practical applications~\cite{auer}.
The MAB is formulated as a mathematical problem without loss of generality and, as such, is related to various stochastic phenomena.
In fact, many application problems in diverse fields, such as communications (cognitive networks~\cite{cog,cog2}), commerce (advertising on the web~\cite{web}), entertainment (Monte-Carlo tree search, which is used for computer games~\cite{uct,mogo}), can be reduced to MABs.

\section{Model}

Kim et al. proposed a MAB solution called ``tug-of-war (TOW),'' which uses a dynamical system.
This algorithm was inspired by the spatiotemporal dynamics of a single-celled amoeboid organism (the true slime mold {\it P. polycephalum})~\cite{kim1,kim2,kim3,kim4,kim5,kim6,kimN}, which maintains a constant intracellular-resource volume while collecting environmental information by concurrently expanding and shrinking its pseudopod-like terminal parts.
In this bio-inspired algorithm, the decision-making function is derived from its underlying physics, which resembles that of a tug-of-war game.
The physical constraint in TOW dynamics, the conservation law for the volume of the amoeboid body, entails a nonlocal correlation among the terminal parts. 
That is, the volume increment in one part is immediately compensated for by volume decrement(s) in the other part(s). 
In our previous studies~\cite{kim1,kim2,kim3,kim4,kim5,kim6,kimN}, we showed that, owing to the nonlocal correlation derived from the volume-conservation law, TOW dynamics exhibit higher performance than other well-known algorithms such as the modified $\epsilon$-greedy algorithm and the modified softmax algorithm, which is comparable to the UCB1-tuned algorithm (seen as the best choice among parameter-free algorithms~\cite{auer}). 
These observations suggest that efficient decision-making devices could be implemented using any physical object as long as it holds some common physical attributes, such as the conservation law.
In fact, Kim et al. demonstrated that optical energy-transfer dynamics between quantum dots, in which energy is conserved, can be exploited for the implementation of TOW dynamics~\cite{QDM,QDM2,QDM3}.

\subsection{ASDM}

\begin{figure}[h]
\centering
\includegraphics[height=125mm]{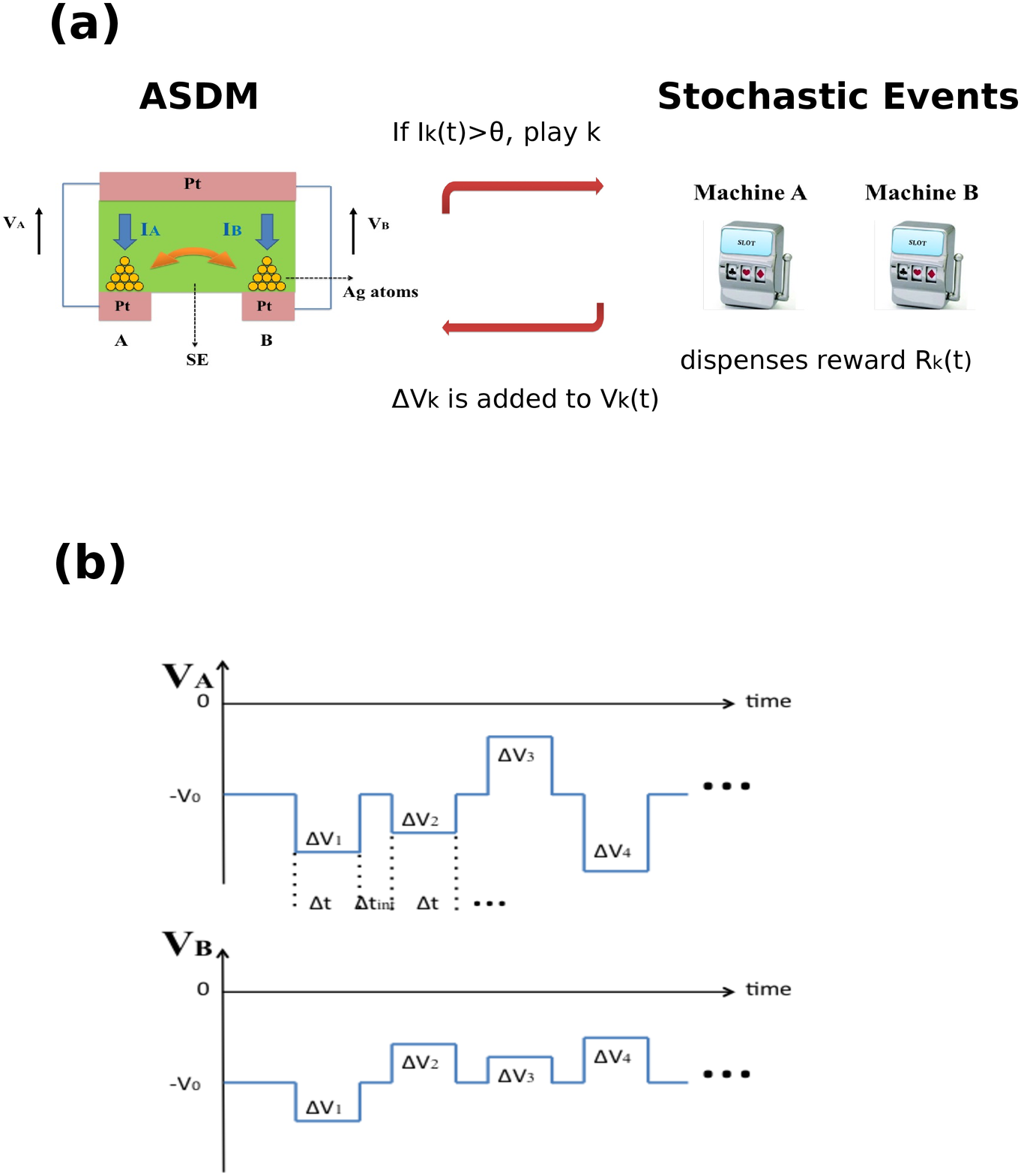}
\caption{(a) The ASDM using gapless type atomic switches. The ASDM decides which machine ($A$ or/and $B$) is to be played at time $t$ according to whether the current $I_k$ is larger than $\theta$ or not. (b) Voltages $V_A$ and $V_B$. Here, added voltage $\Delta V_k (j)$ is determined by each reward $R_k (j)$ (Eq.(\ref{R})) at play $j$ ($I_k$$>$$\theta$).}
\label{fig:stow}
\end{figure}

Here, we propose a simplified model for an atomic switch-based decision maker (ASDM).
Consider two atomic switches located close to each other, in which a solid electrolyte (SE) is sandwiched between one top Pt electrode and two bottom Pt electrodes respectively on both sides, as shown in 
Fig.~\ref{fig:stow}(a).
Each atomic switch is operated in a metal/ionic conductor/metal configuration, which is referred to as a ``gapless type atomic switch~\cite{tsuruoka}.''
Here we assume that operation of both switches is influenced by each other, which implies a certain interaction between the two switches. In the initial state, Ag ions are distributed uniformly in the electrolyte. When a bias voltage of $-V_0$ is applied to the bottom Pt electrodes relative to the top Pt electrode, Ag ions migrate to the bottom electrodes and the same amount of Ag atoms are precipitated on the respective electrode.
We define the height of Ag atoms by $X_0$, and each displacement of the height of precipitated Ag atoms from $X_{0}$ at time $t$ by $X_{k}(t)$ ($k\in \{A,B\}$).
The total height results in $X_0$ $+$ $X_k(t)$.

If current $I_k$$>$$\theta$, we consider that the ASDM chooses machine $k$, and obtains reward $R_k (j)$ generated from each ``unknown PDF'' (mean reward $\mu_k$ is also supposed to be unknown).
According to the reward, the added voltage $\Delta V_k (j)$ is determined by  
\begin{equation}
\Delta V_k (j) =  R_k (j) - K \label{R} .
\end{equation}
Here, $R_k(j)$ is a ``reward'' which has an arbitrary real value.  
$K$ is a parameter to be described in detail later on in this paper.
Then, each voltage becomes 
\begin{equation}
V_k =  - \left( V_0 + \Delta V_k (j) \right) \label{V}.
\end{equation}

We assume the following conditions: 
\begin{enumerate}
\item
At initial equilibrium state, the SE is nearly empty of Ag ions to be precipitated. 
This implies that an increment of one height is compensated by a decrement in the other (Eq.(\ref{diffe}) holds).
\item
If $X_0$ $+$ $X_k(t)$ $>$ $Th$, current $I_k$ is larger than $\theta$.
Here, $Th$ and $\theta$ are thresholds.
If the $Th$ is set to be smaller than $X_0$, this dynamics works from the initial state without fluctuations. 
\item
For simplicity, we assume a linear dependence between $\Delta V_k$ and $\Delta X$ (Eq.(\ref{eq:org})) even though it depends on the shape of the Ag atoms and the amount of Ag ions remaining. 
\item
The time interval for adding voltage $\Delta t$ is sufficiently larger than that for interval $\Delta t_{int}$ that the decaying effect of Ag atoms during $\Delta t_{int}$ can be ignored.  
\end{enumerate}

Displacement $X_A$ ($= - X_B$) is determined by the following equations:
\begin{eqnarray}
X_A(t+1) & = & Q_A(t) - Q_B(t) + \delta(t) ,\label{diffe}\\
Q_k(t) & = & \sum_{j=1}^{N_k}  \Delta V_k (j) . \label{eq:org}
\end{eqnarray}
Here, $Q_k(t)$ ($k\in \{A,B\}$) is an ``estimate'' of information of past experiences accumulated from the initial time $1$ to the current time $t$, $N_k$ counts the number of times that machine $k$ has been played, $\Delta V_k$ is the added voltage when playing machine $k$, $\delta(t)$ is an arbitrary fluctuation to which the body is subjected, and $K$ is a parameter.
Eqs.(\ref{R}) and (\ref{eq:org}) are called the ``learning rule.'' Consequently the ASDM dynamics evolve according to a particularly simple rule: in addition to the fluctuation, if machine $k$ is played at each time $t$, $R_k$$-$$K$ is added to $X_k(t)$ (Fig.~\ref{fig:stow}).

\subsection{SOFTMAX Algorithm}

The SOFTMAX algorithm is a well-known algorithm for solving MAB problems~\cite{sutton}.
In this algorithm, the probability of selecting A or B, $P^{\prime}_A(t)$ or $P^{\prime}_B(t)$, is given by the following Boltzmann distributions:
\begin{eqnarray}
P_A^{\prime}(t) & = &  \frac{\exp[\beta \cdot Q_A(t)]}{\exp[\beta \cdot Q_A(t)] + \exp[\beta \cdot Q_B(t)]},\\
P_B^{\prime}(t) & = &  \frac{\exp[\beta \cdot Q_B(t)]}{\exp[\beta \cdot Q_A(t)] + \exp[\beta \cdot Q_B(t)]},
\end{eqnarray}
where $Q_k(t)$ ($k\in \{A,B\}$) is given by $\frac{\sum_{j=1}^{N_k(t)} R_k(j)}{N_k(t)}$.
Here, $\beta$ is a time-dependent form in our study, as follows: 
\begin{equation}
\beta(t) = \tau \cdot t
\label{eq:beta}.
\end{equation}
$\beta=0$ corresponds to a random selection, and $\beta \rightarrow \infty$ corresponds to a greedy action.

\section{Simulation Results}
\vspace{2mm}

\begin{figure}[h]
\centering
\includegraphics[height=50mm]{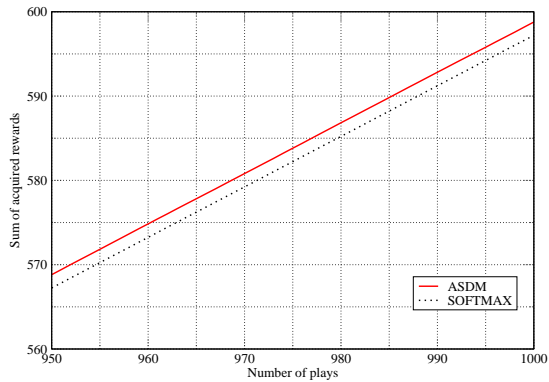}
\vspace{2mm}
\caption{Performance comparison between the ASDM and SOFTMAX. We used $K_0$$=$$0.55$ for the ASDM and $\tau_{opt}$$=$$0.3$ for SOFTMAX.}
\label{fig:sim}
\end{figure}

From computer simulations, we confirmed that, in almost all cases, an ASDM with the parameter $K_0$ ($=$$\frac{\mu_A+\mu_B}{2}$) can acquire more rewards than a SOFTMAX algorithm with the optimized parameter $\tau_{opt}$, although SOFTMAX is well-known as a good algorithm for efficient decision-making~\cite{verm}. 
Figure \ref{fig:sim} shows an example of an ASDM/SOFTMAX performance comparison.
The vertical axis denotes the sum of acquired rewards (mean values of $1000$ samples), and the horizontal axis denotes the number of plays.  
For the reward PDFs, we used normal distributions $N(\mu_A,\sigma^2)$ and $N(\mu_B,\sigma^2)$, where $\mu_A$$=$$0.5$, $\mu_B$$=$$0.6$, and $\sigma$$=$$0.2$.
Computer simulations were executed under the condition that $Th$$=$$X_0$ and $\delta$$=$$sin(\pi/2 + \pi t)$.

\section{Theoretical Analyses of the ASDM}

Theoretical analyses of the TOW dynamics for a Bernoulli type MAB problem, in which a reward is limited to $0$ or $1$, are described in~\cite{kimN}.
In this section, theoretical analyses of the ASDM are described for a general MAB where a reward is not limited to $0$ or $1$.

\subsection{Solvability of the MAB}

\begin{figure}[h]
\centering
\includegraphics[height=70mm]{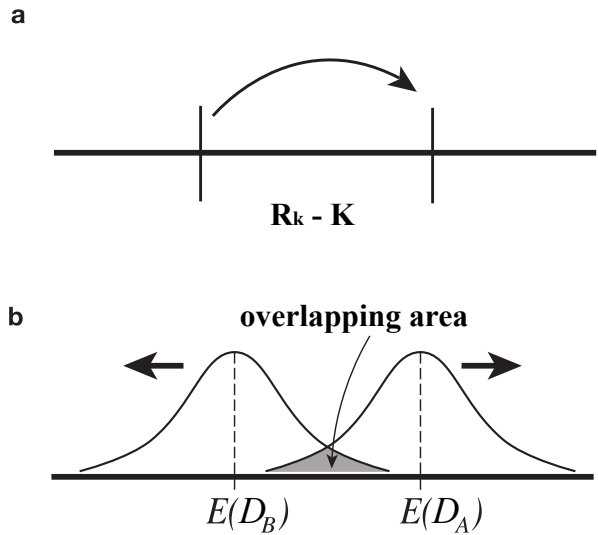}
\caption{(a) Random walk: flight $R_k(t) - K$ . Here, the probability density function of $R_k$ has the mean $\mu_k$. (b) Probability distributions of two random walks.}
\label{fig:random}
\end{figure}

To explore the MAB solvability of the ASDM dynamics, let us consider a random-walk model as shown in Fig.~\ref{fig:random}(a).
Here, $R_k(t)$ ($k\in \{A,B\}$) is a reward at time $t$, and $K$ is a parameter (see Eq.(\ref{R})).
We assume that means of the probability density function of $R_k$ satisfy $\mu_A$ $>$ $\mu_B$ for simplicity. 
After time step $t$, the displacement $D_k(t)$ ($k\in \{A,B\}$) can be described by
\begin{eqnarray}
D_k(t) & = & \sum_{j=1}^{N_k(t)} R_k(j) - K \hspace{1mm}N_k(t) . \label{eq:ran} 
\end{eqnarray}
The expected value of $D_k$ can be obtained from the following equation:
\begin{equation}
E( D_k(t) ) = (\mu_k - K )\hspace{1mm} N_k(t) .
\end{equation}

In the overlapping area between the two distributions shown in Fig.~\ref{fig:random}(b), we cannot accurately estimate which is larger.
The overlapping area should decrease as $N_k$ increases so as to avoid incorrect judgments.  
This requirement can be expressed by the following forms:
\begin{eqnarray}
  \mu_A - K      & > & 0 ,  \\
  \mu_B - K     & < & 0 . \label{eq:cond1}
\end{eqnarray}
These expressions can be rearranged into the form  
\begin{equation}
\mu_B  < K   < \mu_A .  \label{eq:cond1}
\end{equation}
In other words, the parameter $K$ must satisfy the above conditions so that the random walk correctly represents the larger judgment.

We can easily confirm that the following form satisfies the above conditions:
\begin{equation}
K   = \frac{\mu_A + \mu_B}{2} .\label{eq:fixed}
\end{equation}
From $Q_k(t)$ and Eq.(\ref{eq:fixed}), we obtain
\begin{eqnarray}
K_0 & = & \frac{\gamma}{2}  , \label{eq:w0}\\
\gamma & = & \mu_A + \mu_B .
\end{eqnarray}
Here, we have set the parameter $K$ to $K_0$.
Therefore, we can conclude that the ASDM dynamics using the learning rule $Q_k$ with the parameter $K_0$ can solve the MAB correctly.

\subsection{Origin of the high performance}
 
In many popular algorithms such as the $\epsilon$-greedy algorithm, at each time $t$, an estimate of reward probability is updated for either of the two machines being played.
On the other hand, in an imaginary circumstance in which the sum of the mean rewards $\gamma$ $=$ $\mu_A$ $+$ $\mu_B$ is known to the player, we can update both of the two estimates simultaneously, even though only one of the machines was played.

\begin{table}[h]
\caption{Estimates for each mean reward based on the knowledge that machine $A$ was played $N_A$ times and that machine $B$ was played $N_B$ times---on the assumption that the sum of the mean rewards $\gamma$ $=$ $\mu_A$ $+$ $\mu_B$ is known.}
\label{table:1}
\begin{center}
\begin{tabular}{|c|c|c|c|}\hline \hline
$A$: & {$\frac{\sum_{j=1}^{N_A} R_A(j)}{N_A}$} & $B$: & $\gamma \hspace{1mm} -$ {$\frac{\sum_{j=1}^{N_A} R_A(j)}{N_A}$} \\ \hline
$A$: & $\gamma \hspace{1mm} -$ {$\frac{\sum_{j=1}^{N_B} R_B(j)}{N_B}$} & $B$: & {$\frac{\sum_{j=1}^{N_B} R_B(j)}{N_B}$} \\ \hline
\end{tabular}
\end{center}
\end{table}
The top and bottom rows of Table~\ref{table:1} provide estimates based on the knowledge that machine $A$ was played $N_A$ times and that machine $B$ was played $N_B$ times, respectively.
Note that we can also update the estimate of the machine that was not played, owing to the given $\gamma$.

From the above estimates, each expected reward $Q^{\prime}_k$ ($k\in \{A,B\}$) is given as follows:
\begin{eqnarray}
Q^{\prime}_A & = & N_A \hspace{1mm} \frac{\sum_{j=1}^{N_A} R_A(j)}{N_A} + N_B \hspace{1mm} \bigl( \gamma \hspace{1mm} - \frac{\sum_{j=1}^{N_B} R_B(j)}{N_B} \bigr) \nonumber \\
 & = & \sum_{j=1}^{N_A} R_A(j) - \sum_{j=1}^{N_B} R_B(j) + \gamma N_B, \label{eq:qAp}\\
Q^{\prime}_B & = & N_A\hspace{1mm} \bigl( \gamma \hspace{1mm} - \frac{\sum_{j=1}^{N_A} R_A(j)}{N_A} \bigr) + N_B\hspace{1mm} \frac{\sum_{j=1}^{N_B} R_B(j)}{N_B} \nonumber \\
& = & \sum_{j=1}^{N_B} R_B(j) - \sum_{j=1}^{N_A} R_A(j) + \gamma \hspace{1mm} N_A. \label{eq:qBp}
\end{eqnarray}
These expected rewards, $Q^{\prime}_j$s, are not the same as those given by the learning rules of TOW dynamics, $Q_j$s in Eqs.(\ref{R}) and (\ref{eq:org}).  
However, what we use substantially in TOW dynamics is the difference 
\begin{equation}
Q_A - Q_B  =  (\sum_{j=1}^{N_A} R_A(j) - \sum_{j=1}^{N_B} R_B(j)) - K \hspace{1mm}(N_A - N_B)\label{eq:dq}.
\end{equation}
When we transform the expected rewards $Q^{\prime}_j$s into $Q^{\prime \prime}_j  =  Q^{\prime}_j / 2 $, 
we can obtain the difference
\begin{equation}
Q^{\prime \prime}_A - Q^{\prime \prime}_B  =  (\sum_{j=1}^{N_A} R_A(j) - \sum_{j=1}^{N_B} R_B(j)) - \frac{\gamma}{2} \hspace{1mm} (N_A - N_B). \label{eq:dqpp}
\end{equation}
Comparing the coefficients of Eqs.(\ref{eq:dq}) and (\ref{eq:dqpp}), the differences in their constituent terms are always equal when $K = K_0$ (Eq.(\ref{eq:w0})) is satisfied.
Eventually, we can obtain the nearly optimal weighting parameter $K_0$ in terms of $\gamma$.

This derivation implies that the learning rule for the ASDM dynamics is equivalent to that of the imaginary system in which both of the two estimates can be updated simultaneously.
In other words, the ASDM dynamics imitates the imaginary system that determines its next move at time $t+1$ in referring to the estimates of the two machines, even if one of them was not actually played at time $t$. 
This unique feature in the learning rule, derived from the fact that the sum of mean rewards is given in advance, may be one of the origins of the high performance of the ASDM dynamics.

Monte Carlo simulations were performed it was verified that the ASDM dynamics with $K_0$ exhibits an exceptionally high performance, which is comparable to its peak performance---achieved with the optimal parameter $K_{opt}$.
To derive the optimal value $K_{opt}$ accurately, we need to take into account the fluctuations.

In addition, the essence of the process described here can be generalized to $M$-machine cases.
To separate distributions of the top $m$-th and top $(m+1)$-th machine, as shown in Fig.~\ref{fig:random}(b), all we need is the following $K_0$:
\begin{eqnarray}
K_0  & = & \frac{\gamma^{\prime}}{2} , \\
\gamma^{\prime} & = & \mu_{(m)} + \mu_{(m+1)}
\end{eqnarray} 
Here, $\mu_{(m)}$ denotes the top $m$-th mean, and $m$ is any integer from 1 to
$M - 1$. The MBP is a special case where $m = 1$.
In fact, for $M$-machine and $X$-player cases, we have designed a physical system that can determine the overall optimal state, called the ``social maximum,'' quickly and accurately~\cite{bom,bom2}.

\subsection{Performance characteristics}

To characterize the high performance of the ASDM dynamics, let us consider the imaginary model for solving the MAB, called the ``cheater algorithm.''
The cheater algorithm selects a machine to play according to the following estimate $S_k$ ($k\in \{A,B\}$)
\begin{eqnarray}
S_A  & = & X_{A, 1} + X_{A, 2}, + \cdots + X_{A, N} , \\
S_B  & = & X_{B, 1} + X_{B, 2}, + \cdots + X_{B, N} .
\end{eqnarray} 
Here, $X_{k, i}$ is a random variable.
If $S_A$ $>$ $S_B$ at time $t=N$, machine $A$ is played at time $t=N+1$. 
If $S_B$ $>$ $S_A$ at time $t=N$, machine $B$ is played at time $t=N+1$. 
If $S_A$ $=$ $S_B$ at time $t=N$, a machine is played randomly at time $t=N+1$. 
Note that the algorithm refers to results of both machines at time $t$ without any attention to which machine was played at time $t-1$.
In other words, the algorithm ``cheats'' because it plays both machines and collects both results, but declares that it plays only one machine at a time.   

The expected value and the variance of $X_k$ are defined as $E (X_k)  =  \mu_k $ and $V (X_k)  =  \sigma_k^{2}$.
Here, $\mu_k$ is the same as the $P_k$ defined earlier.
From the central-limit theorem, $S_k$ has a Gaussian distribution with $E (S_k)  =  \mu_k N$ and $V (S_k)  =  \sigma_k^{2} N$. 
If we define a new variable $S = S_A - S_B$, 
$S$ has a Gaussian distribution and carries the following values:
\begin{eqnarray}
E (S)  & = & (\mu_A + \mu_B) N  ,\\
V (S)  & = & (\sigma_A^{2} +  \sigma_B^{2}) N ,\\
\sigma (S)  & = & \sqrt{ \sigma_A^{2} +  \sigma_B^{2}} \sqrt{N} .
\end{eqnarray} 

\begin{figure}[h]
\centering
\includegraphics[height=30mm]{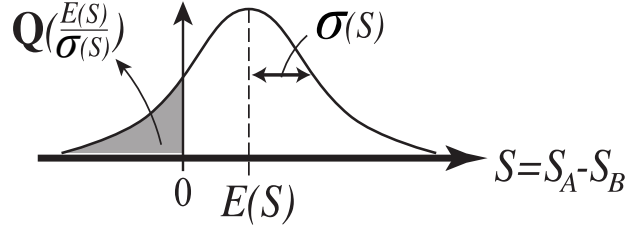}
\caption{{\bf Q}$(\frac{E(S)}{\sigma(S)})$: probability of selecting the lower-reward machine in the cheater algorithm}
\label{fig:Qfunc}
\end{figure}
 
From Fig.~\ref{fig:Qfunc}, the probability of playing machine $B$, which has a lower reward probability, can be described as {\bf Q}$(\frac{E(S)}{\sigma(S)})$.
Here, {\bf Q}$(x)$ is a {\bf Q}-function.
We obtain 
\begin{eqnarray}
P(t=N+1,B) & = & {\bf Q}(\phi \sqrt{N}) .
\end{eqnarray}
Here, 
\begin{eqnarray}
\phi & = & \frac{\mu_A - \mu_B}{\sqrt{\sigma_A^2 + \sigma_B^2}} .
\end{eqnarray}

Using the Chernoff bound ${\bf Q}(x)  \leq  \frac{1}{2} \exp(-\frac{x^2}{2})$, 
we can calculate the upper bound of a measure, called the ``regret,'' which quantifies the accumulated losses of the cheater algorithm.
\begin{equation}
regret = (\mu_A - \mu_B) E(N_B).
\end{equation}
\begin{eqnarray}
E(N_B) & = & \sum_{t=0}^{N-1} {\bf Q}(\phi \sqrt{t}) \nonumber \\
       & \leq & \sum_{t=0}^{N-1} \frac{1}{2} \exp(-\frac{\phi^2}{2} t) \nonumber \\
       & = & \frac{1}{2} + \sum_{t=1}^{N-1} \frac{1}{2} \exp(-\frac{\phi^2}{2} t) \nonumber \\
       & \leq & \frac{1}{2} + \int_{0}^{N-1} \frac{1}{2} \exp(-\frac{\phi^2}{2} t) dt \nonumber \\
       & = & \frac{1}{2} - \frac{1}{\phi^2} \left( \exp(-\frac{\phi^2}{2}(N-1))  - 1  \right) \\
       & \rightarrow & \frac{1}{2} + \frac{1}{\phi^2} .
\end{eqnarray} 
Note that the regret becomes constant as $N$ increases.

Using the ``cheated'' results, we can also calculate the regret of the ASDM dynamics in the same way.
In this case, 
\begin{eqnarray}
S_A  & = & X_{A, 1} + X_{A, 2}, + \cdots + X_{A, N_A} - K N_A , \\
S_B  & = & X_{B, 1} + X_{B, 2}, + \cdots + X_{B, N_B} - K N_B .
\end{eqnarray} 
$X_{k, i}$ is also a random variable.
Then, we obtain 
\begin{eqnarray}
E (S_k)  & = & ( \mu_k - K )  N_k ,\\
V (S_k)  & = & \sigma_k^{2} N_k . 
\end{eqnarray} 
Using the new variables $S = S_A - S_B$, $N = N_A + N_N$, and $D = N_A - N_N$, we also obtain 
\begin{eqnarray}
E (S)  & = & \frac{\mu_A - \mu_B}{2} N + \left( \frac{\mu_A + \mu_B}{2} - K \right) D   ,\\
V (S)  & = & \frac{\sigma_A^2 + \sigma_B^2}{2} N +  \frac{\sigma_A^2 - \sigma_B^2}{2} D .
\end{eqnarray}

If the conditions $K  = K_0$ and $\sigma_A = \sigma_B$ $\equiv \sigma$ are satisfied, 
we then obtain 
\begin{eqnarray}
E (S)  & = & \frac{\mu_A - \mu_B}{2} N ,\\
V (S)  & = & {\sigma^2} N ,
\end{eqnarray} 
and 
\begin{eqnarray}
P(t=N+1,B) & = & {\bf Q}(\phi_{T} \sqrt{N}) .   
\end{eqnarray}
Here, 
\begin{eqnarray}
\phi_{T} & = & \frac{\mu_A - \mu_B }{2 \sigma} .
\end{eqnarray} 

We can then calculate the upper bound of the regret for the ASDM dynamics
\begin{eqnarray}
E(N_B) & = & \sum_{t=0}^{N-1} {\bf Q}(\phi_{T} \sqrt{t}) \nonumber \\
       & \leq & \frac{1}{2} - \frac{1}{\phi_{T}^2} \left( \exp(-\frac{\phi_{T}^2}{2}(N-1))  - 1  \right) \\
       & \rightarrow & \frac{1}{2} + \frac{1}{\phi_{T}^2} .
\end{eqnarray} 
Note that the regret for the ASDM dynamics also becomes constant as $N$ increases.

It is known that optimal algorithms for the MAB, defined by Auer et al., have a regret proportional to $\log(N)$~\cite{auer}. 
The regret has no finite upper bound as $N$ increases because it continues to require playing the lower-reward machine to ensure that the probability of incorrect judgment goes to zero.  
A constant regret means that the probability of incorrect judgment remains non-zero in the ASDM dynamics, although this probability is nearly equal to zero.
However, it would appear that the reward probabilities change frequently in actual decision-making situations, and their long-term behavior is not crucial for many practical purposes.
For this reason, the ASDM dynamics would be more suited to real-world applications.

\section{Conclusion and Discussion}

In this paper, we proposed an ASDM for solving MAB problems, and analytically validated that their high efficiency in making a series of decisions for maximizing the total sum of stochastically obtained rewards is embedded in a volume-conserving physical system when subjected to suitable operations involving fluctuations.
In conventional decision-making algorithms for solving MAB problems, the parameter for adjusting the ``exploration time'' must be optimized.
This exploration parameter often reflects the difference between the rewarded experiences, i.e., $|\mu_A - \mu_B |$.
In contrast, the ASDM demonstrates that higher performance can be achieved by introducing a parameter $K_0$ that refers to the sum of the rewarded experiences, i.e., $\mu_A$ $+$ $\mu_B$.
This type of optimization, using the sum of the rewarded experiences, is particularly useful for time varying environments (reward probability or reward PDF)~\cite{kim2}.  
Owing to this novelty, the high performance of the TOW dynamics can be reproduced when implementing these dynamics with atomic switches.

The ASDM proposed in this paper is a simple ``ideal model.''
While the assumptions used for constructing the model may contain some points that do not match real experimental situations, we can more accurately extend the model so that the modified assumptions do match real experimental situations. 
As long as the TOW dynamics between atomic switches is implemented, high performance decision-making can be guaranteed even in the extended model. 

The ASDM will introduce a new physics-based analog computing paradigm, which will include such things as ``intelligent nanodevices'' and ``intelligent information networks'' based on self-detection and self-judgment.
Thus, our proposed physics-based analog-computing paradigm would be useful for a variety of real-world applications and for understanding the biological information-processing principles that exploit their underlying physics.

\section*{Acknowledgments}

The authors thank Dr. Etsushi Nameda and Prof. Masashi Aono for valuable discussions in the early stages of this work. 
This work was supported in part by the Sekisui Chemical Grant Program for ``Research on Manufacturing Based on Innovations Inspired by Nature.''

\section*{Conflict of Interest}

The authors declare that there is no conflicting of interest regarding the publication of this paper.


\end{document}